\documentclass{article}

\usepackage{microtype}
\usepackage{graphicx}
\usepackage{subfigure}
\usepackage{booktabs} 
\usepackage{enumitem}
\usepackage{multirow}
\usepackage{amssymb}
\usepackage{pdfrender}
\usepackage{subfigure}
\usepackage{array}

\usepackage{hyperref}

\newcommand{\1}{\textcircled{\scriptsize{1}}}
\newcommand{\2}{\textcircled{\scriptsize{2}}}

\usepackage[accepted]{mlsys2022}

\mlsystitlerunning{Prediction of GPU Failures Under Deep Learning Workloads}

\begin{document}

\twocolumn[
\mlsystitle{Prediction of GPU Failures Under Deep Learning Workloads}



\mlsyssetsymbol{equal}{*}

\begin{mlsysauthorlist}
\mlsysauthor{Heting Liu}{psu}
\mlsysauthor{Zhichao Li}{bytedance}
\mlsysauthor{Cheng Tan}{neu}
\mlsysauthor{Rongqiu Yang}{bytedance}
\mlsysauthor{Guohong Cao}{psu}
\mlsysauthor{Zherui Liu}{bytedance}
\mlsysauthor{Chuanxiong Guo}{bytedance}
\end{mlsysauthorlist}

\mlsysaffiliation{psu}{
The Pennsylvania State University}
\mlsysaffiliation{bytedance}{ByteDance}
\mlsysaffiliation{neu}{
Northeastern University}

\mlsyscorrespondingauthor{Heting Liu}{hxl476@psu.edu}

\mlsyskeywords{Machine Learning, GPU failures}

\vskip 0.3in

\begin{abstract}
Graphics processing units (GPUs) are
the de facto standard for processing deep learning (DL) tasks.
Meanwhile,
GPU failures, which are inevitable, cause severe consequences
in DL tasks:
they disrupt distributed trainings,
crash inference services,
and result in service level agreement violations.
To mitigate the problem caused by GPU failures,
we propose to predict failures by using ML models.

This paper is the first to study prediction models of GPU failures
under large-scale production deep learning workloads.
As a starting point, we evaluate classic prediction models and
observe that predictions of these models are both inaccurate and unstable.
To improve the precision and stability of predictions,
we propose several techniques,
including parallel and cascade model-ensemble mechanisms
and a sliding training method.
We evaluate the performances of our various techniques on a four-month production dataset including 350 million entries.
The results show that our proposed techniques improve the
prediction precision from
46.3\% to 84.0\%.
\end{abstract}
]



\printAffiliationsAndNotice{}

\section{Introduction}
\label{sec:introduction}

In a large-scale deep learning cluster, GPU failures are inevitable,
and they cause severe consequences.
For example,
a failure on one GPU can disrupt a long-running distributed training job.
Developers lose multiple hours of work (sometimes even tens of hours) on multiple machines.
Furthermore, GPU failures may
crash production inference services,
increase responding latency significantly,
causes service level agreement (SLA) violations,
and result in revenue loss.

GPU failure prediction is critical to avoid the revenue loss.
For example, measures such as proactive maintenance and rescheduling can be taken to reduce the loss, if knowing the failure information beforehand. 
However, GPU failure prediction under DL workloads is not well-studied.
Some related works~\cite{tiwari2015understanding, nie2016large, nie2017characterizing, nie2018machine, gupta2015understanding}
are analyzing GPU failures on Titan supercomputer, a high performance computing (HPC) system.
These studies are inspiring, but they provide limited guidance to predict today's GPU failures in large-scale DL clusters because (i) DL workloads are considerably different from HPC workloads, (ii) the GPU type being studied---NVIDIA K20X GPU---is about a decade old (launched in 2012), and (iii) None of them comprehensively studied how to predict the GPU failures.  
Beyond GPU failures,
people have also examined other hardware failures in
DRAM~\cite{sridharan2013feng, meza2015revisiting},
disks~\cite{pinheiro2007failure},
SSDs~\cite{schroeder2016flash},
co-processors~\cite{oliveira2017experimental},
and other datacenter hardware~\cite{vishwanath2010characterizing}.
These studies have the same spirit as ours
in studying failures,
but our study focuses on the unique characteristics
of GPUs, and we aim to predict GPU failures under DL workloads.
This paper is the first to study the prediction of GPU failures under production-scale GPU clusters.

To study the prediction of GPU failures, we deploy a system that constantly collects data from running GPUs.
For model training purpose, we convert the GPU dataset to a time series dataset and adopt a sliding-window approach
to improve the number of positive instances in the training dataset.
As a starting point, we explore various classic models such as long short-term memory (LSTM) and one-dimensional
convolutional neural network (1D-CNN), and evaluate their performances for GPU failure prediction.
We observe that predictions of these models are both inaccurate and unstable (e.g., decreasing over time).

We learn that the GPU failure pattern is complex and diverse, and the learning ability of one single model may be limited.
Therefore, we propose two model-ensemble mechanisms (i.e., parallel and cascade) that use multiple models, to make combined
predictions. The parallel model-ensemble mechanism uses different models to further verify the prediction results.
The cascade model-ensemble mechanism first filters out 95\% ``healthy'' instances, and then combine with the latter model
to make predictions.

We also observe that the pattern of GPU failure may change over time, making the trained model less effective.
To adapt the model to the changing patterns, we propose a sliding training technique.
Specifically, we retrain the model periodically, using only recently collected data.
We further observe that the time length of the sliding training set affects the model performance, and yet there is no
one-to-all optimal time length of the sliding training set. Therefore, we also propose to adjust the time length of
the sliding training set, in a dynamic manner, to better train the model.

The contributions of this paper are summarized as follows:
\begin{itemize}
\vspace{-0.05in}
\item This is the first to study prediction models of GPU failures under large-scale production deep learning workloads.
\vspace{-0.05in}
\item We identify two major challenges for these prediction models: precision and stability.
\vspace{-0.05in}
\item We propose several techniques, including parallel and cascade model-ensemble mechanisms, and a sliding training method,
	to improve the precision and stability of the prediction models.
\end{itemize}

We evaluate the model performances on a four-month dataset including about 350 million entries.
The precision of the best baseline model (i.e., 1D-CNN) is only 46.3\%. Our proposed model-ensemble mechanism
is able to improve the precision by up to 11.9\% (refer to Section \ref{sec:eval_ensemble}).
The sliding training technique improves the stability of the model, and also improves the average precision by 22.8\% compared
to the one without it (refer to Section \ref{sec:eval_slide}).
The training set length-adjustment technique can further improve the precision by up to 5.7\%.
We achieve the highest average precision of 84.0\%, with all techniques combined.

\section{Background}
\label{sec:background}
In this section, we introduce GPU deployments at CompanyXYZ, elaborate the GPU data we collect and how we generate the time series dataset for model training.

\subsection{GPU Deployment and Workload}

An important factor that influences GPU failures
is how GPUs are deployed and used.
In this section, we briefly introduce the GPU setup 
and their workloads at CompanyXYZ.
CompanyXYZ has multiple datacenters worldwide with tens of thousands of GPUs.
GPUs are organized in an 8-card-per-machine setup.
Machines are connected through RDMA-enabled datacenter networks.
We collect data from GPUs serving DL model training and inference.
The training jobs in CompanyXYZ are large-scale, some of which (for example, training GPT-3)
require a collaboration of almost a thousand GPUs.
Our inference services are also massive,
as we need to process billions of model inference requests per day.

In this paper, we sample a subset of CompanyXYZ's GPUs,
and we focus on three recent GPU types: V100, P4, and T4.
The data are collected from tens of thousands of running GPUs across multiple datacenters in Asia.
In the rest of this paper, we will refer these data (all of them) as \textit{raw GPU dataset}, or \textit{GPU dataset},
which we describe in details later on.
The GPUs and machines are chosen at random,
hence our observations ought to be general.

\begin{table*}[t]
\centering
\small
\begin{tabular}{@{\extracolsep{4pt}}llll@{}cll}
\hline
\multirow{2}{*}{Parameters} & \multicolumn{3}{c}{Dataset} & \multicolumn{3}{c}{Failure Prediction} \\
                               \cline{2-4}                     \cline{5-7}
                             & Type & Range & Source      & Used & DType & Usage \\
\hline
\multicolumn{7}{l}{dynamic data}\\
\hline
temperature         & int & $(20, 90)$ (in $ ^\circ C$) & \texttt{nvidia-smi}              & \checkmark & float  & feature \\
power consumption   & int & $[0, 400)$ (in W)           & \texttt{nvidia-smi}              & \checkmark & float  & feature \\
GPU SM utilization  & int & $[0, 100]$                  & \texttt{nvidia-smi}              & \checkmark        & --     & feature \\
GPU mem utilization & int & $[0, 100]$                  & \texttt{nvidia-smi}              & \checkmark & float  & feature \\
machine uptime      & int & --                          & \texttt{/proc/}                  & \checkmark & float  & feature \\
failure status      & bit & --                   & failure report  & \checkmark & binary & label \\
\hline
\multicolumn{7}{l}{static data}\\
\hline
IP address          & string     & --                           & \texttt{ifconfig}        & -- & categorical & -- \\
GPU serial number   & string     & --                           & cmdb   & --         & --          & ID\\
machine rack name   & string     & --                           & cmdb                     & \checkmark & categorical & feature \\
GPU Position        & int        & $[0, 7]$                     & cmdb                     & --         & --          & --\\
GPU type            & string     & \{``V100'', ``T4'', ``P4''\} & cmdb                     & \checkmark & categorical & feature \\
driver version      & string     & \{``418'', ``450''\}         & cmdb                     & \checkmark & categorical & feature \\
expiration date     & date       & --                           & machine management sys   & \checkmark & float       & feature \\
\hline
\end{tabular}
\caption{GPU and machine parameters in the GPU dataset and their usage in our prediction model.
Note that ``Type'' in the ``GPU Dataset'' is the data type from underlying sources (e.g., \texttt{nvidia-smi}),
whereas ``DType'' in the ``Failure Prediction'' is the type we use for ML model training.
Other than data types, we convert ``failure status'' into binary to represent if a GPU is faulty,
and we cut ``machine rack name'' to datacenter names for training.
}
\label{tab:dataset}
\end{table*}

\subsection{Raw GPU Dataset}

We collect both static and dynamic GPU attributes.
Table \ref{tab:dataset} lists the description of each attribute. Most parameters are self-explanatory. One parameter, ``failure status'' requires some explanation.
 The ``failure status'' indicates whether the GPU is failed or not at this time point, where ``1'' means the GPU is failed and ``0'' means healthy.
The GPU failure in this paper is defined from a user's perspective, that is the errors are classified based on their \textit{externally observable consequences}, regardless of their internal causes. This observation-based definition aligns with our goal, that is, to predict failures which affect services and applications. 

\subsection{Data Preprocessing}
\label{sec:data_prepare}

\begin{figure*} [t]
   \centering 
   \subfigure[Negative Instance]{ 
     \label{fig:neg_instance}  
     \includegraphics[width=2.8in]{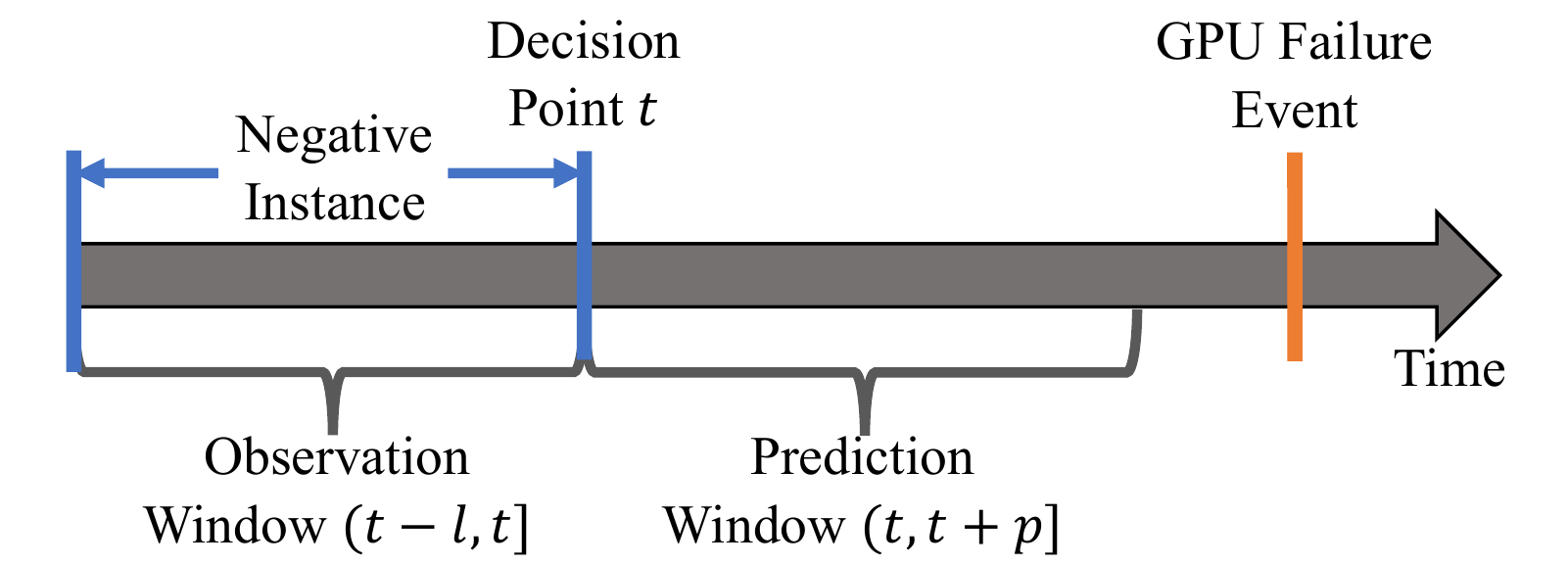}} 
   \hspace{0.1 in} 
   \subfigure[Positive Instance]{ 
     \label{fig:pos_instance} 
     \includegraphics[width=2.8in]{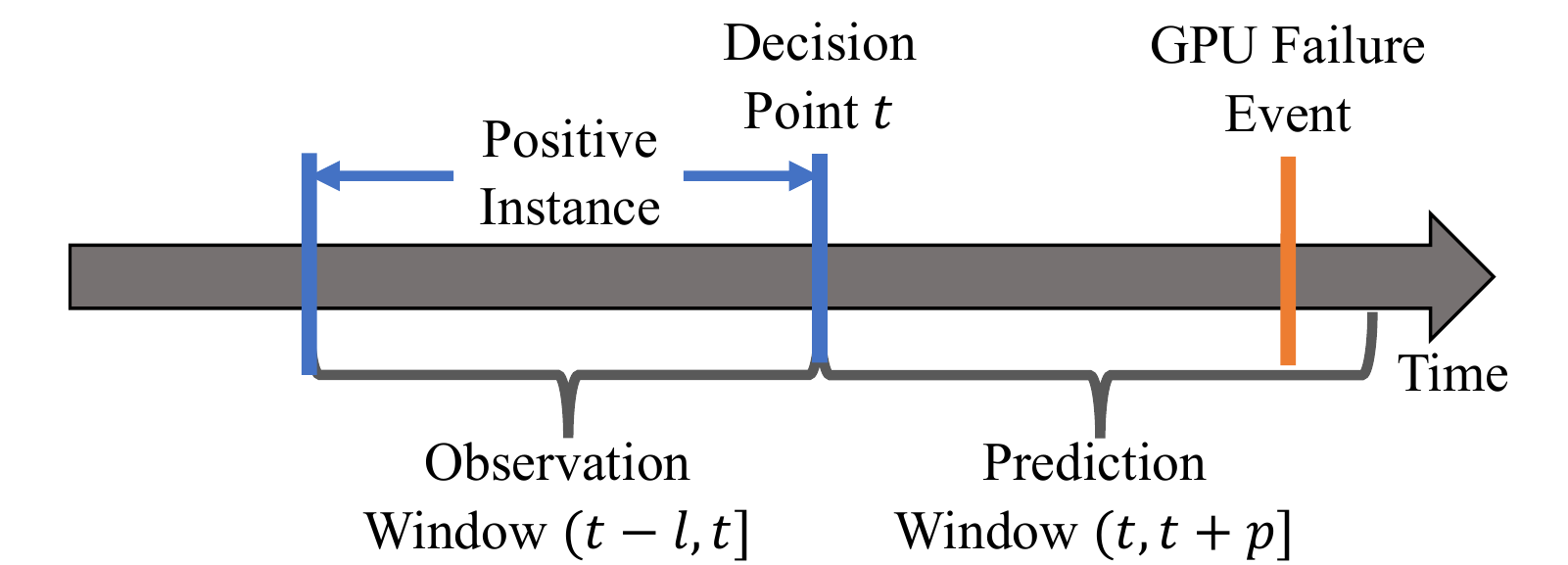}} 
   \caption{Examples of time-series instances generation. (a) A negative instance is generated if the GPU will not fail within time $p$. (b) A positive instance is generated if the GPU will fail within time $p$}
   \label{fig:instances} 
 \end{figure*}

In this paper, we aim to predict whether a GPU will fail within certain time period, e.g., one day, in the future.
After collecting the raw GPU dataset, we generate a time series dataset to train the prediction model.
Formally, let $D$ denote the time-series dataset. $D=\{(\mathbf{X}^{i}, y^{i})\}_{i=1}^{N}$, where $N$ is the
number of instances in the $D$.
$\mathbf{X}^{i}$ denotes a time series, with the size of $l \times m$, where $l$ is the number of time steps in
one instance, and $m$ is the dimension of the feature vector.
Let $p$ denote the prediction length. $y^{i}$ is an indicator variable with $y^{i}=1$ meaning the GPU will
fail within time $p$, and $y^{i}=0$ meaning the GPU will not fail within time $p$.

Next, we elaborate how we convert the raw GPU dataset to dataset $D$.
First, we aggregate the entries of each GPU by GPU serial number. The entries of each GPU are sorted by time
and the interval between two consecutive entries is 10 minutes. Since a failed GPU may not be repaired immediately,
once a GPU fails, its status will be maintained as ``1'' until it is repaired. We filter out the redundant
entries, i.e., removing failure status being ``1'' following the first failure.
Each time-series instance consists of $l$ number of consecutive entries, corresponds to $l$ time steps.
Suppose the timestamp of the last entry in $\mathbf{X}^{i}$ is $t$, the label of the instance depends on the failure
status of the following entries during time $(t, t+p]$. If the failure status is 0 for all the entries during time $(t, t+p]$,
the label of this time series is 0, meaning that the corresponding GPU will not fail within time $(t, t+p]$. If there is
any entry with failure status 1 during $(t, t+p]$, the label of this time series is 1, meaning that the corresponding GPU
will fail within that time period.

To generate the time-series instances, we first explore a segmented approach as described below.
The idea is to split the raw data (i.e., a long time series for each GPU) into different non-overlapping time series instances.
As illustrated in Figure~\ref{fig:instances}, for each GPU,
an observation window of length $l$ (including $l$ time steps) starts from the initial time point.
We first check whether the failure status of all entries in the window is 0.
If it is not, we move the observation window right until there is no failure status being ``1'' in the window.
Then the data in the current observation window forms a time series instance, where each entry corresponds to one row
of $\mathbf{X}^{i}$. Suppose the timestamp of the last entry in the window is $t$, we further check the failure status
of entries between timestamp $t$ and $t+p$: if there are any entries with failure status ``1'', the label $y^{i}$ of
the current time series instance is 1, otherwise the label is 0.
After that, we move the observation window to the entry right after the prediction window, and repeat the steps
above to generate the following time series instances.

However, the number of positive instances produced by the segmented approach is quite small, making training
less effective. To augment the positive instances, we propose a sliding-window approach: after one time-series
instance is generated, the observation window slides $slide\_step$ ($slide\_step < l$) entries to generate the
next time series instance.
One failure event generates multiple positive time-series instances this way.
The number of the positive instances is improved by 60 more times compared to that of the segmented
sampling, when $l=18$,  $p=144$, and $slide\_step = 10$.


\section{Staring Point: Classic Models}
\label{sec:baseline}

In this section, we introduce the classic models we explore and the main challenges we observe from the results.

\subsection{Feature Engineering}
Before training models, we encode the collected features into 
fixed-length numerical feature vectors.
As shown in Table~\ref{tab:dataset}, the type of features we collected includes binary, categorical, and float.
For categorical features such as ``gpu type'' and ``gpu version'', we use one-hot encoding. Each category is first converted to an integer $n$, indicating that it belongs to the $n^{th}$ category. Then we encode it into a one-hot vector, whose dimension is the number of categories. The $n^{th}$ element of the one-hot vector is 1, and other elements are 0.
For float features, the feature value is discretized into $N_{bucket}$ buckets and converted to a bucket index. Such conversion reduces the influences of extreme values for model training.

\subsection{Classic Models}

As a starting point, we build some classic machine learning models and evaluate their performances on GPU failure prediction.
Our problem can be seen as a binary classification problem, that is to classify a time series as ``healthy'' or ``failure''. Commonly used methods for binary
classification problems are Gradient Boosting Decision Tree (GBDT), Multi-layer Perception (MLP), Long Short-Term Memory (LSTM), and etc. We consider the
following models:

\textbf{GBDT}: an ensemble prediction model which combines multiple weak models. 
At each iteration, a new decision is trained with respect to the gradient of the loss achieved by the previous decision trees. 
Our GBDT model ensembles 200 decision trees. 

\textbf{MLP}: a class of feedforward artificial neural network designed to approximate any continuous function and solve problems which are not linearly separable. We build a MLP with two hidden layers.

\textbf{LSTM}: an artificial recurrent neural network (RNN) architecture capable of learning order dependence in sequence prediction problems. We build a LSTM model with hidden state size to be 10.

\textbf{1D-CNN}: a kind of CNN whose kernel moves in 1 direction. 1D-CNN is mostly used on time series data. We build a 1D-CNN model with four convolutional layers and two fully-connected layers.

We train the above models using data collected from March 16th to 31th, 2021, and test the performances of models on the following 6 days. 
The model output for each instance is a score, denoted as $\hat{p}^i$, which is a float number that indicates the probability that the instance belongs to the failure class.
We focus on the instances with the top K predicted scores since they are the most likely positive (failure class) instances.
Specifically, we rank all instances by their predicted score $\hat{p}^i$ from high to low,
then we evaluate the performances of models with \textbf{Precision@K}~\cite{jarvelin2017ir} which corresponds to the ratio of true positive instances among the
total top K instances (ranked by scores). In the rest of this paper, we use precision@K and precision interchangeably.

\begin{figure}[t!]
 \centering
   \includegraphics[scale=0.48]{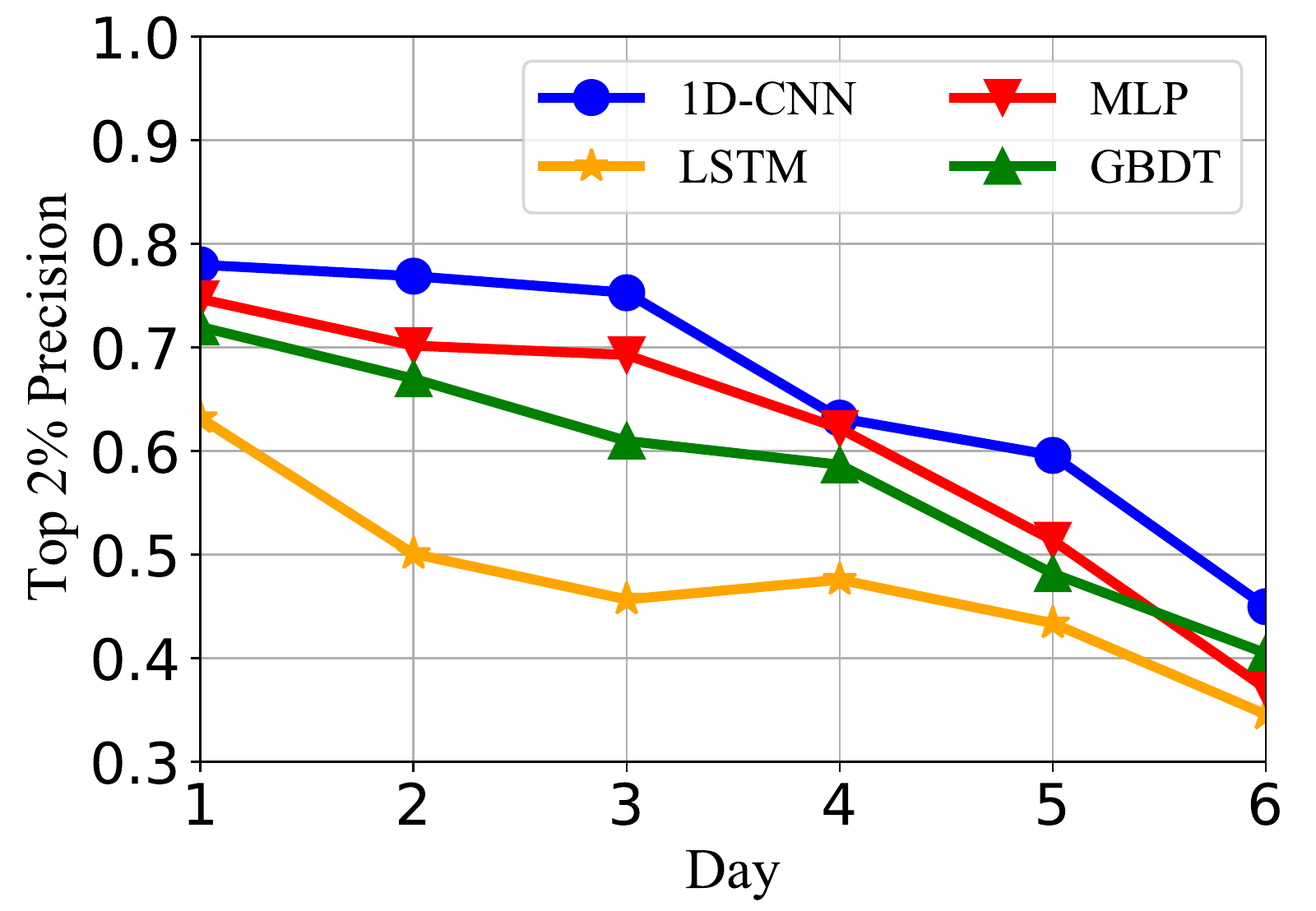}
   \caption{Precision@K of GBDT, MLP, LSTM and 1D-CNN (K=2\%) of each day. The y-axis starts at 0.3.}
   \label{fig:precision_baseline}
 \end{figure}

Figure~\ref{fig:precision_baseline} shows the daily precision@K of the above models. Here we set K=2\% to focus on the most likely positive instances.
From the figure we observe that 1D-CNN model achieves the highest precision@K among the four models. However, the precision@K of 1D-CNN is still quite low, which is 0.663 on average. The average precision@K of GBDT, MLP and LSTM is even lower, which are 0.579, 0.608, 0.474, respectively. 
From Figure~\ref{fig:precision_baseline} we also observe that the precision of all the models decreases as time goes by. The precision decreases by 28.6\%-43.0\% from the first day to the sixth day for the four models.  

From the above results, we hypothesize that \1 the GPU failure pattern is complex and diverse, and thus the overall precisions of the classic models are low. 
\2 The GPU failure pattern may change and thus the model precision may decrease over time. 
Based on these hypotheses, we propose two model-ensemble mechanisms in Section~\ref{sec:ensemble} to improve the precision, and a sliding training technique in
Section~\ref{sec:slide_train} to improve the stability issue of predictions.

\section{Model-Ensemble Technique}
\label{sec:ensemble}

The pattern of GPU failures is complex and diverse, and therefore, one single classic model may not capture the pattern well.
To improve the precision of GPU failure prediction, we propose two model-ensemble mechanisms, namely parallel and cascade, to ensemble multiple models in
different manners, to make combined predictions more effectively.

\subsection{Parallel}
\label{sec:parallel}

\begin{figure}[t!]
 \centering
    \includegraphics[scale=0.58]{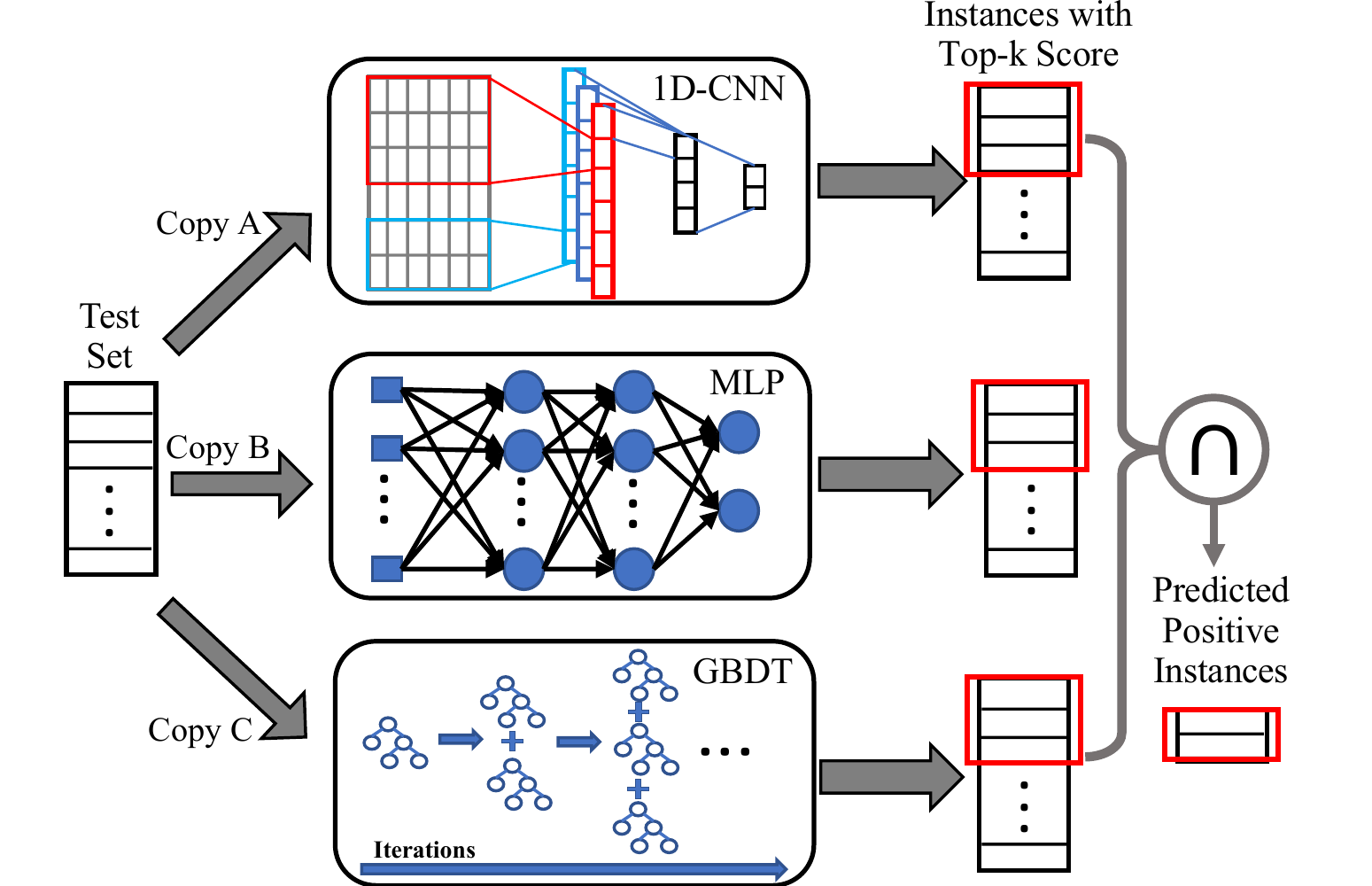}
    \caption{The structure of the parallel mechanism. The figure shows the inference procedure when using the parallel mechanism. Each instance is fed into the three models respectively, and the instance ranked top K by three models is predicted to be the ``failure'' class.}
   \label{fig:parallel}
\end{figure}

Figure~\ref{fig:parallel} shows the inference procedure of the parallel mechanism. 
We use three different models to make joint decisions. 
Specifically, the instances in the test set are fed into the 1D-CNN, MLP, and GBDT respectively, and each model selects a set of instances ranked top K according to its predicted scores. 
Then the instances in the intersection of the three sets is classified as the ``failure'' class. 
For the purpose of capturing different failure patterns, we exploit three different models for GPU failure prediction.
We take the intersection of the top K sets output by three models because the instance is more likely to be accurate if all models predict it to fail soon, and
thus the precision of prediction can be improved.

\subsection{Cascade}
\label{sec:cascade}

\begin{figure*}[t!]
 \centering
   \includegraphics[scale=0.61]{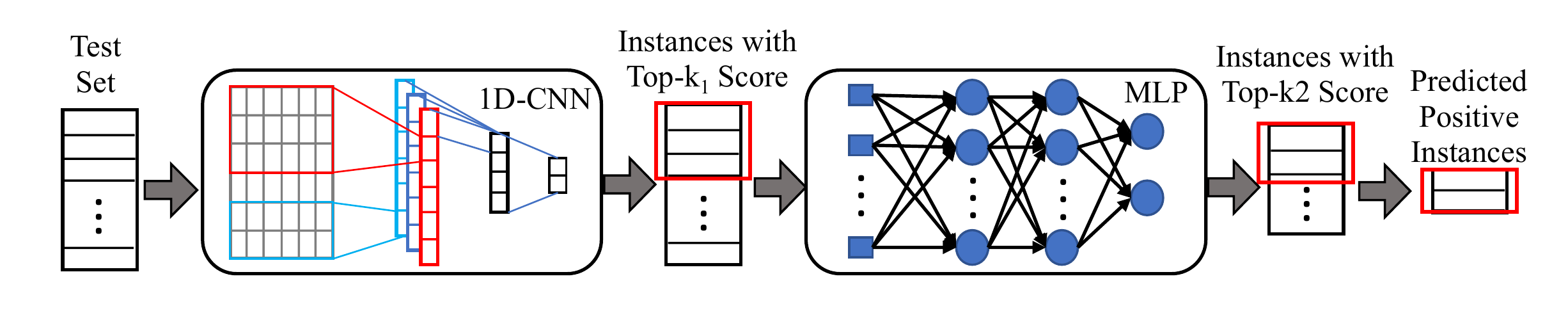}
   \caption{The structure of the cascade mechanism. The figure shows the inference procedure when using the cascade mechanism. The instances in the test set are
	first fed into the 1D-CNN model to filter out some negative instances, and then the latter MLP model further classifies the instances.}
   \label{fig:cascade}
 \end{figure*}

The second ensemble approach is a cascade mechanism, as shown in Figure~\ref{fig:cascade}.
We choose 1D-CNN model and MLP model to do the cascade procedure because these two models have the highest
two precisions as shown in Section~\ref{sec:baseline}. 
The idea of this mechanism is to first filter out instances that are most likely ``healthy'', and then use the
second model to further classify the instances.

As illustrate in Figure~\ref{fig:cascade}, the test set instances are first fed into the 1D-CNN model to filter out the most likely ``healthy'' instances. 
To achieve this, we set a weight $w^i$ for each instance to control the penalty in case of classifying it incorrectly when training the 1D-CNN model.
We set the weight of the positive instances to be much higher than that of the negative instances, so that the punishment of
classifying positive instances to be negative is high.
Let $g_{\theta}(\cdot)$ denote the 1D-CNN model with parameters $\theta$, then the loss function of the 1D-CNN model is
\begin{equation}
\label{eq:weighted_loss}
    loss = \frac{1}{N}\sum_{i=1}^N | w^i (g_{\theta}(\mathbf{X}^{i}) - y^{i}) |^2.
\end{equation}
By minimizing Eq.(\ref{eq:weighted_loss}), the predicted scores of positive instances tend to be high. 
Then we sort the instances by their predicted score, and select the top $k_1$ ranked instances. Most negative instances will be filtered out this way.
We select the top $k_1$ instances, which should include most positive instances and some of the negative instances, and feed them into the second model (i.e., MLP model) for further classification. 
Finally the top $k_2$ ranked instances predicted by the MLP model is classified as the ``failure'' ones.

\section{Sliding Training Technique}
\label{sec:slide_train}

In Section~\ref{sec:baseline} we observe that the precision of GPU failure prediction decreases over time, probably due to the fact that the GPU failure pattern
changes over time. This could be caused by many factors: GPU driver upgrade, humidity and illumination in the environment, etc.

\begin{figure}[ht]
\centering
   \includegraphics[scale=0.53]{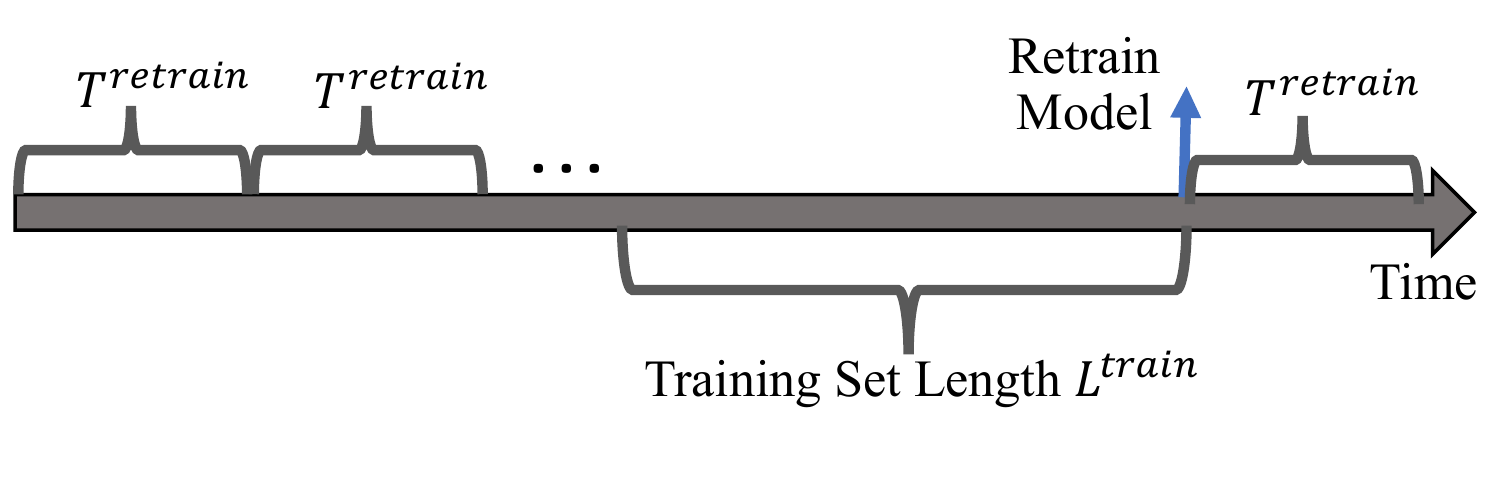}
   \caption{The procedure of sliding training. $T^{retrain}$ denotes the period of model retraining, and $L^{train}$ denotes the length of the time span of training set. The model is retrained every $T^{retrain}$, using the data of previous $L^{train}$ days as the training data.}
   \label{fig:retrain}
\end{figure}

To cope with the pattern changing problem, we propose to slide the training set to the recently collected data, and retrain the model periodically so that the current pattern can be captured.  
As illustrate in Figure~\ref{fig:retrain}, let $T^{retrain}$ denote the model retrain period, and $L^{train}$ denote the training set time length. The model is retained at $n \times T^{retrain}, where n=1,2,\cdots$, and
the training set is generated from the data collected during $[n T^{retrain}-L^{train}, n T^{retrain}]$. Each $[n T^{retrain}, (n+1) T^{retrain}]$ is called a test window. 
The model is updated to match the recent data this way.

As mentioned above, the speed of the pattern changes may not be constant. Thus different value of $L^{train}$ may produce different effects at different time.   
To explore the effect of $L^{train}$,
we trained three models with $L^{train}$ to be 9 days, 12 days and 15 days for each test window, and test their performances. 
Table~\ref{tab:Ltrain1} and Table~\ref{tab:Ltrain2} show two examples of the performance of the parallel model at different time. 
From the tables we see that for test window April 10th-12th, setting $L^{train}$ to be 9 days has the highest precision@K, while for test window April 13th-15th, setting $L^{train}$ to be 12 days has the highest precision@K.
The results verify that at different time, the optimal $L^{train}$ may be different. 

Generally longer $L^{train}$ includes more data, and thus model is more likely to learn the mapping well. But it may not be sensitive to the changing pattern since the data may
include many stale patterns. Shorter $L^{train}$ only includes the most recent data which better help capture the recent pattern, and is 
sensitive to the changing patterns. However, it has the risk of overfitting. Therefore, 
when the pattern changing is smooth, longer $L^{train}$ may work better, while when the pattern changing is rapid, smaller $L^{train}$ may be better. 

From the above analysis, we can see that there is no one-to-all optimal $L^{train}$. The approach to automatically adjust $L^{train}$ to achieve higher
precision has yet to be studied thoroughly in our future work.
Currently, we manually adjust $L^{train}$, which we referred to as variable-length sliding training method. For example, for the test window April 10th-12th,
2021, the model is trained with $L^{train}$ being 9 days. And for the test window April 13th-15th, 2021, the model is trained with $L^{train}$ being 12 days. 
One potential solution to automatically adjust $L^{train}$ is that we train multiple models with different values of $L^{train}$ each time we retrain the model, and the $L^{train}$ that achieves the highest precision on the previous test window is selected to be used for the current test window. 
Other potential methods such as an AutoML based approach are also discussed in Section~\ref{sec:discuss}.

\begin{table}[tb]
  \centering
  \begin{tabular}{l c c c}
    \hline
     $L^{train}$ & Precision@K & Recall@K & Accuracy \\ \hline
     15 days & 77.5\% & 12.9\% & 89.9\%\\ 
     12 days & 79.6\% & 10.3\% & 89.7\% \\ 
     \textbf{9 days} & \textbf{88.1\%} & \textbf{11.5\%} & \textbf{90.0\%} \\ \hline   
  \end{tabular}
     \caption{The precision@K of the parallel mechanism under different $L^{train}$s on test window April 10th-12th.}
   \label{tab:Ltrain1}
\end{table}

\begin{table}[tb]
  \centering
  \begin{tabular}{l c c c}
    \hline
     $L^{train}$ & Precision@K & Recall@K & Accuracy \\ \hline
     15 days & 60\% & 2.1\% & 88.7\%\\ 
     \textbf{12 days} & \textbf{67.4\%} & \textbf{6.3\%} & \textbf{89.3\%} \\ 
     9 days & 53.0\% & 1.6\% & 88.7\%\\ \hline
  \end{tabular}
   \caption{The precision@K of the parallel mechanism under different $L^{train}$s on test window April 13th-15th.}
   \label{tab:Ltrain2}
\end{table}

\section{Implementation}
\label{sec:implement}

\begin{figure}[t]
\begin{center}
\includegraphics[width=0.48\textwidth]{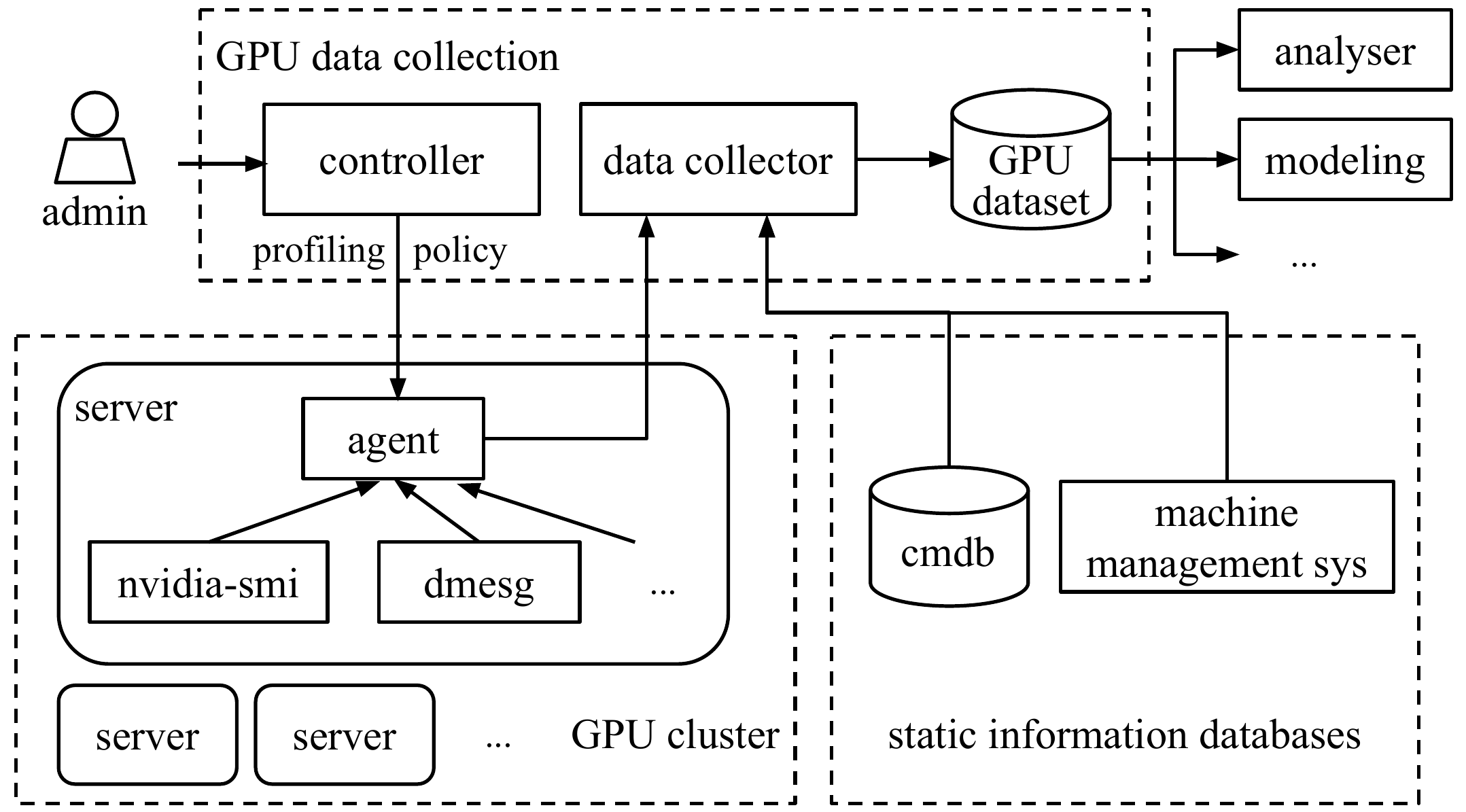}
\end{center}
\caption{GPU data collection infrastructure.
  ``cmdb'' means configuration management database,
  a standard database to store information about hardware assets.
  ``machine management sys'' is a CompanyXYZ internal system that tracks
  machine-level information (e.g., purchased date, expiration date).
}
\label{fig:arch}
\end{figure}

In this section, we introduce our data collection infrastructure to collect the raw GPU dataset.
At CompanyXYZ,
we have a data collecting system
that constantly fetching running status from GPUs both when they are healthy and
encountering failures.
It also gathers GPUs' static configuration information, for example,
expiration dates and which rack a GPU locates.
We build and deploy a data collection infrastructure 
based on existing tools (e.g., \texttt{nvidia-smi}, \texttt{dmesg}, service management systems at CompanyXYZ).
It periodically collects runtime data from GPUs,
combines runtime data with static configurations,
and updates the raw GPU dataset.
Figure~\ref{fig:arch} depicts the architecture of this data collection
infrastructure.

This system works as follows.
First, an administrator decides a \textit{collecting policy} which specifies
what data to collect and how frequently the data is collected,
and send this policy to a \textit{controller}.
The controller then broadcasts
this policy to data collecting \textit{agents} (a daemon process) running on servers.
Agents are responsible for collecting data from different
underlying data sources---for example, \texttt{nvidia-smi}, \texttt{dmesg},
and \texttt{/proc/}---periodically.
Agents are also in charge of failure detection.
If a failure is detected, the agent will record the failure context and send a failure report to
CompanyXYZ's failure handling system (omitted in Figure~\ref{fig:arch}).

\textit{Data collector} receives data from agents, including both normal runtime data and failure reports.
The collector combines these dynamic data
with static configuration data which is stored in other CompanyXYZ's management databases.
For example,
configuration management database (\textit{cmdb})
is one such database that contains machine and GPU hardware information.
Data collector joins all these data by GPU serial number, a unique identifier
for each GPU, and updates the~GPU~dataset.

Finally, the updated GPU dataset is stored on HDFS and is used to generate the time-series dataset as
described in Section~\ref{sec:data_prepare} for model training purpose.
Our model training is implemented using Tensowflow2.0 in Python.

\section{Evaluation}
\label{sec:eval}

In this section, we evaluate the proposed techniques respectively. Specifically, we answer the following questions:
\begin{itemize}
    \vspace{-0.05in}
    \item What is the performance of the parallel mechanism and the cascade mechanism, and how do they compare to baselines?
    \vspace{-0.05in}
    \item How to set the retrain period in sliding training?
    \vspace{-0.05in}
    \item How much do sliding training and variable-length sliding training help improve the precision and stability of predictions?
\end{itemize}

\textbf{Experiment setup.}
We collect four-month production data from March, 2021 to June, 2021 to generate the time series dataset.
We evaluate the models based on parallel mechanism (referred to as parallel model) and cascade mechanism (referred to as cascade model) against several baseline models, i.e., 1D-CNN, MLP, and GBDT. 
We choose these baseline models because they are the basic components in the ensemble mechanisms.
We evaluate the models with the following metrics:
\begin{itemize}
    \item Precision@K as introduced in Section~\ref{sec:baseline}. In the evaluations, we set K to be 2\% as explained in Section~\ref{sec:baseline}.
    \vspace{-0.1in}
    \item Recall@K~\cite{malheiros2012source}:
    Recall at K is the ratio of true positive instances within top K instances (ranked by score) among the total positive instances:
    \vspace{-0.1in}
    \begin{equation}
        Recall@K = \frac{\sum_{i=1}^K y^i}{\sum_{i=1}^N y^i}.
    \end{equation}
    Similar to precision@K, we set K to be 2\%.
    \vspace{-0.1in}
    \item Accuracy. The fraction of predictions the model gets right. We set the $threshold$ of prediction score to be 0.7, i.e., an instance $i$ is predicted to be positive if $\hat{p}^i>threshold$, and negative otherwise.
\end{itemize}

\textbf{Data Balancing.}
Since the number of negative instances is much more than that of the positive instances, the training set is highly imbalanced,
and may result in poor performance of models, especially for the minority class (``failure'' class). Therefore, we under sample
the negative instances and over sample the positive instances when training the models, to make the training set balanced.
The ratio of the positive and negative instances in the training set after sampling is set to 1:1, a commonly used ratio when
balancing dataset for model training purpose.
On the other side, for test purpose, the ratio of the positive and negative instances in the test set is set to as large as 1:8.

\subsection{Evaluation of Ensemble Mechanisms}
\label{sec:eval_ensemble}

\begin{table}[tb]
  \centering
   \scalebox{1.0}{
  \begin{tabular}{l c c c}
    \hline
    Model & Precision@K & Recall@K & Accuracy \\ \hline 
    1D-CNN &  46.3\% & 8.3\% & 87.6\%  \\ 
    MLP &  44.5\% & 7.9\% & 86.8\%  \\ 
    GBDT &  42.3\% & 7.6\% & 87.0\%  \\ \hline
    Parallel &  58.2\% & 7.1\% & 89.3\%  \\ 
    Cascade &  50.1\% &  8.0\% & 89.1\%  \\ \hline 
  \end{tabular}
  }
  \vspace{-0.1in}
    \caption{Comparison of parallel model, cascade model, 1D-CNN, MLP and GBDT on data in April.}
  \label{tab:ensemble_perf}
\end{table}

In this section, we answer the first question.
We first evaluate parallel and cascade models against baselines to validate the effectiveness of the two ensemble mechanisms. 
Then we evaluate the stability of these models by testing the daily precision@K.
All models in this section are trained without sliding training.

We train the above models using data collected from March 16th to 31th, 2021, and test the performances of models
using data collected from April 1st to 30th, 2021.
Table~\ref{tab:ensemble_perf} presents the results of 1D-CNN, MLP, GBDT, parallel and cascade model.
From the table we observe that both parallel model and cascade model achieve higher precision@K than all the baseline models. 
The parallel model improves precision@K from 46.3\% (achieved by the best baseline 1D-CNN) to 58.2\%, and the cascade model improves the precision@K from 46.3\% to 50.1\%.
Similar results are obtained on data in May and June.

 \begin{figure}[t!]
 \centering
   \includegraphics[scale=0.44]{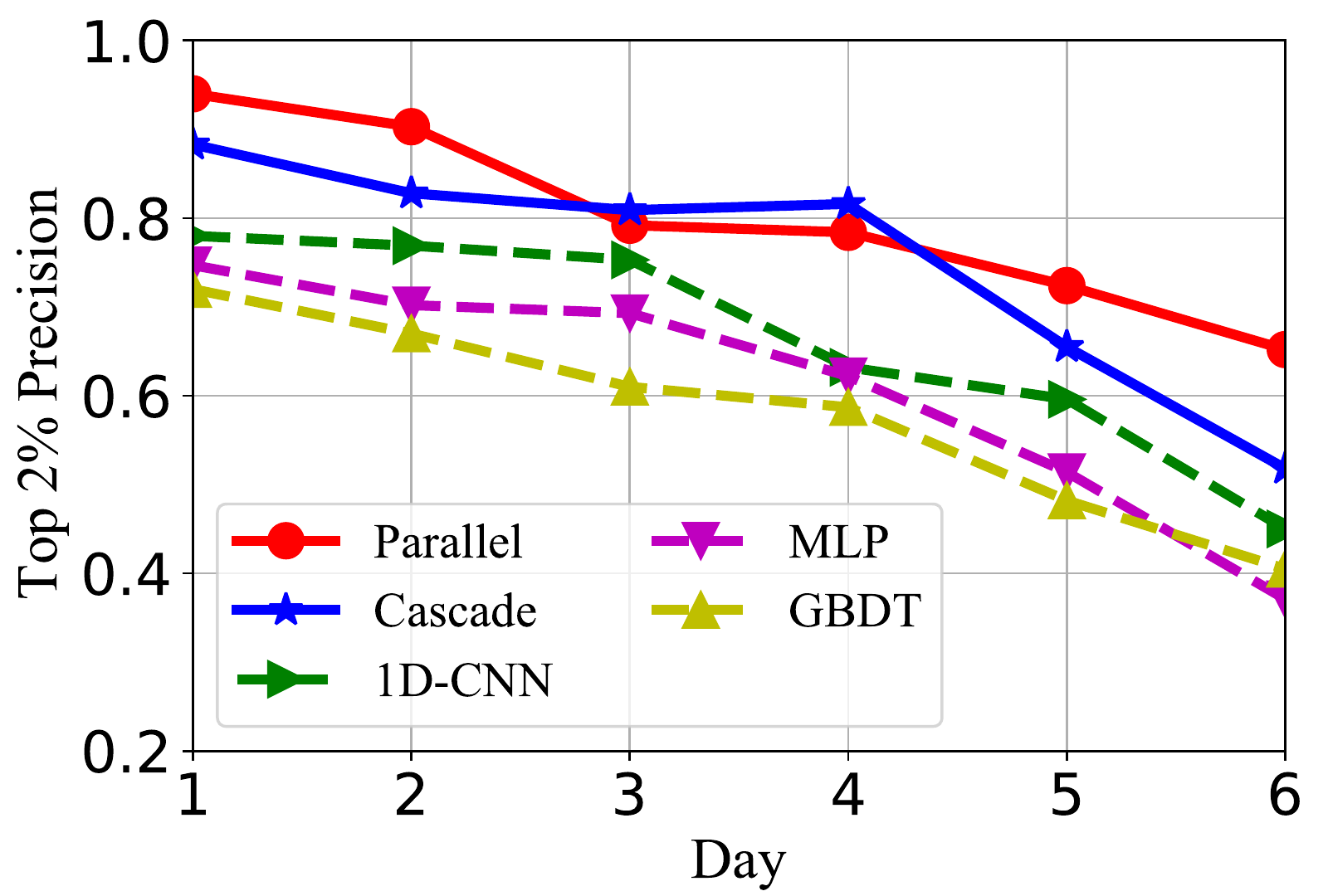}
   \caption{Precision@K comparison of parallel model, cascade model, 1D-CNN, MLP and GBDT. The y-axis starts at 0.2.}
   \label{fig:ensemble_perf}
 \end{figure}
 
To evaluate the stability of predictions, we further test the daily precision@K. 
Figure~\ref{fig:ensemble_perf} shows the precision@K from April 1st to 6th as an example.
From the figure we observe that both the parallel model and the cascade model outperform all baselines all the time.
But similar to the baseline models, their precision@K decreases with time.

\subsection{Evaluation of Retrain Period}

\begin{figure}[t!]
 \centering
   \includegraphics[scale=0.44]{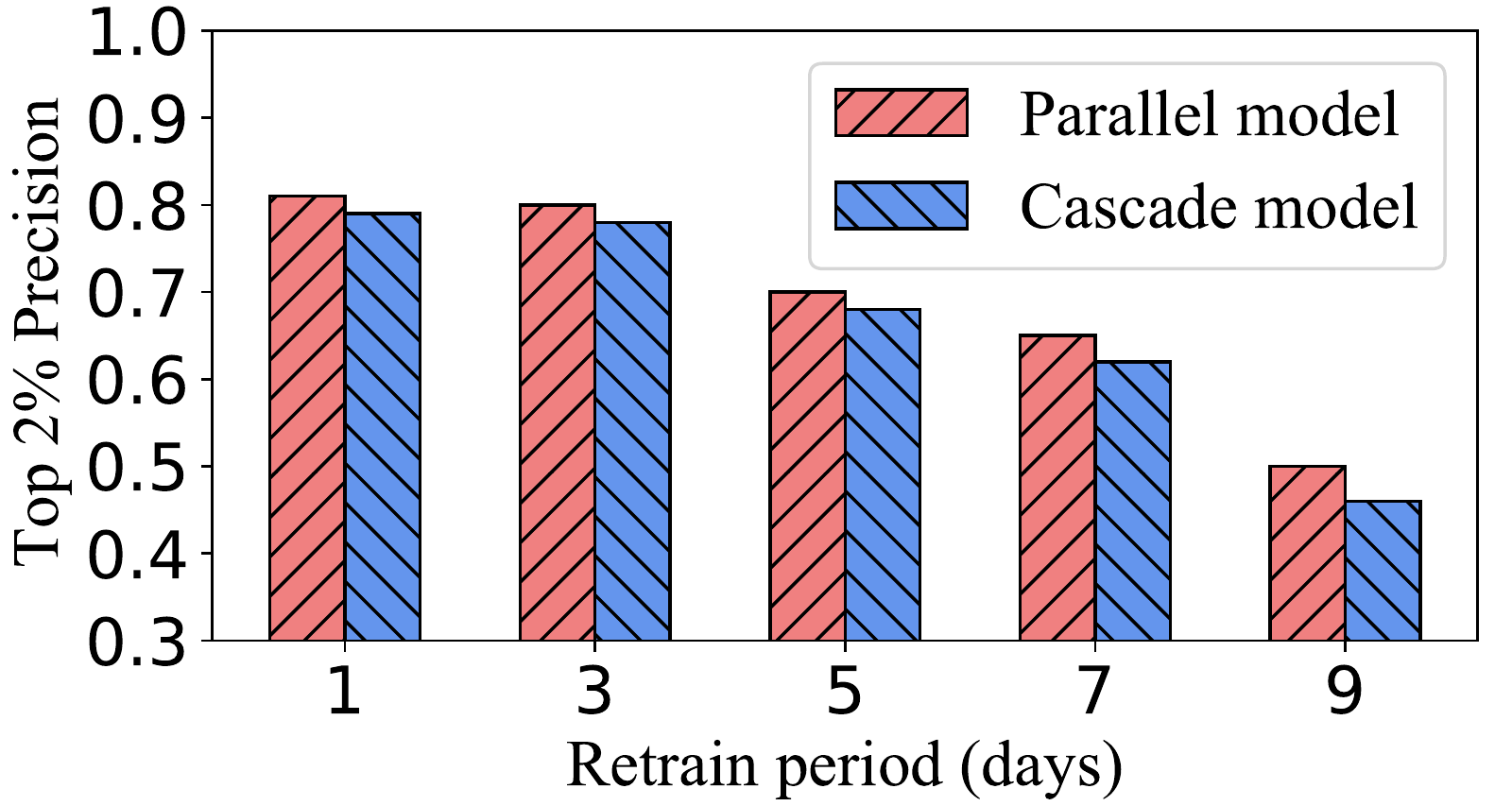}
   \caption{Precision@K v.s. retrain period of parallel model and cascade model. The y-axis starts at 0.3.}
   \label{fig:retrain_freq}
   \vspace{-0.5cm}
\end{figure}

In this section, we answer the second question: how we determine the retrain period, that is how often we should retrain the model in sliding training.
We set $L^{train}$ to be 15 days, and evaluate the models under different retrain periods. Specifically, we set the retrain period $T^{retrain}$ (defined in Section~\ref{sec:slide_train}) to be 1 day, 3 days, 5 days, 7 days and 9 days, respectively, and test the corresponding model performances from April 1st to 30th. 

Figure~\ref{fig:retrain_freq} shows the average precision@K in April when we retrain the model under different $T^{retrain}$. 
From the figure we observe that the average precision@K when $T^{retrain}$ is set to be 3 days is similar to that when $T^{retrain}$ is set to be 1 day.
However, when $T^{retrain}$ is set to be longer (i.e., 5 days, 7 days and 9 days), the precision@K significantly drops. 
Similar results are obtained in on data in May and June.
Therefore, we set the retrain period to be 3 days in the following experiments.

\subsection{Evaluation of Sliding Training}
\label{sec:eval_slide}

In this section, we evaluate how much the sliding training helps improve the precision and stability of predictions. The training set length $L^{train}$ is set to be 15 days for sliding training.
Table~\ref{tab:slide_train} shows the performance from April 1st to 30th of the parallel model and the cascade model with and without sliding training, respectively. 
From the table we observe that with sliding training, the overall performances of both parallel model and cascade model are significantly improved. 
Especially, the precision@K is improved from 58.2\% to 80.0\% for the parallel model, and from 50.1\% to 78.1\% for the cascade model. The accuracy and recall@K are also improved for both two models with sliding training.

\begin{table}[tb]
  \centering
 \begin{tabular}{l @{ }r  @{ \space\space}r  @{ \space\space}r}
    \hline
    Model & Precision@K & Recall@K & Accuracy \\ \hline
    Parallel (NS) & 58.2\% & 7.1\% & 89.3\%  \\   
    Parallel (Sliding) & 80.0\% & 10.0\% & 89.8\%  \\
    Cascade (NS) & 50.1\% &  8.0\% & 89.1\%   \\
    Cascade (Sliding) & 78.1\% & 14.1\% & 90.3\%  \\  \hline
  \end{tabular}
  \vspace{-0.1in}
    \caption{Comparison of parallel model and cascade model, with sliding training (Sliding) and without sliding training (NS) on data in April.}
  \label{tab:slide_train}
\end{table}

\begin{figure}[t!]
     \centering
     \subfigure[Parallel model.]{
         \includegraphics[width=1.57in]{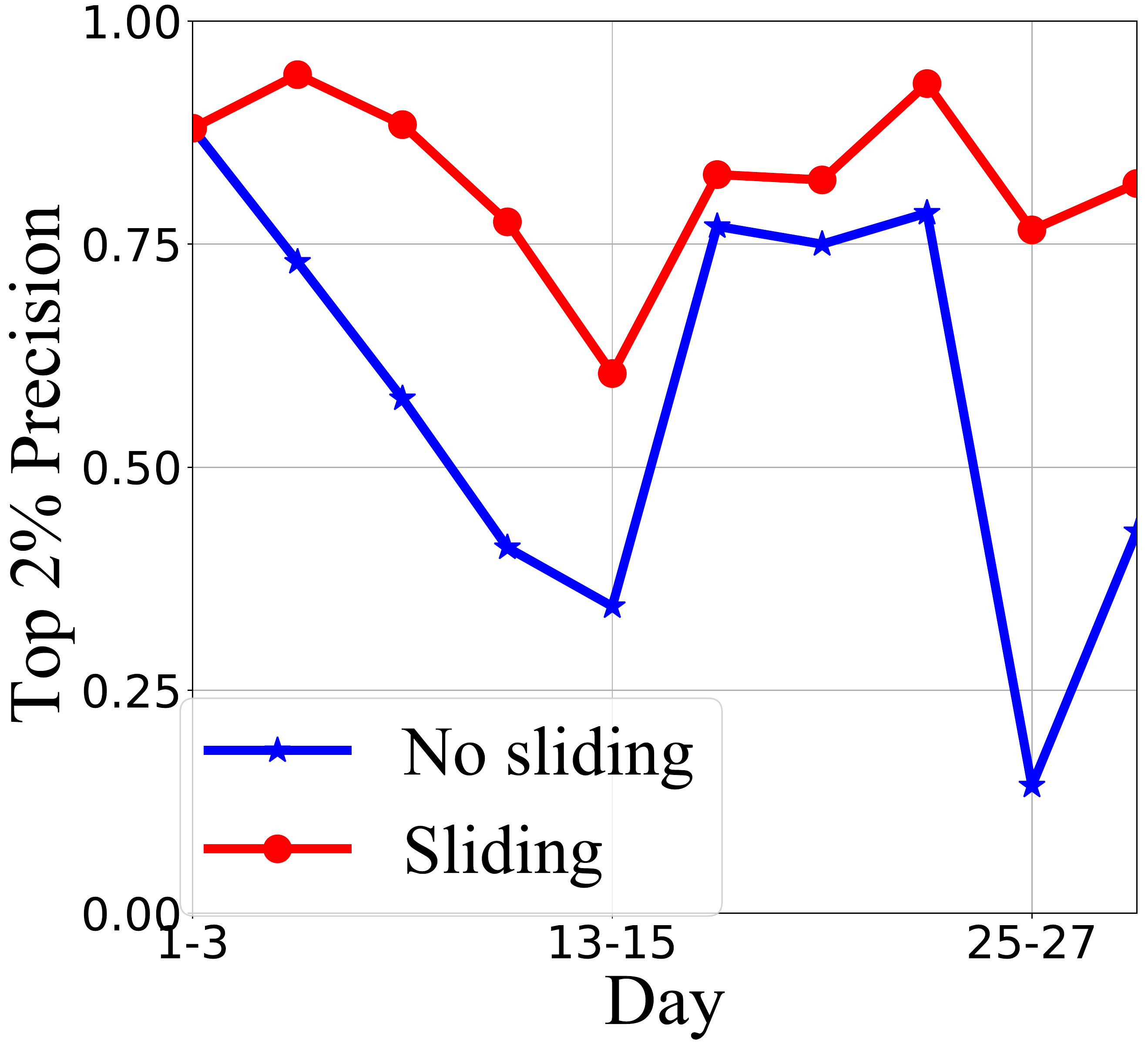}
     \label{fig:para_slide}
     }
    \hspace{-0.1 in} 
     \subfigure[Cascade model.]{
         \includegraphics[width=1.57in]{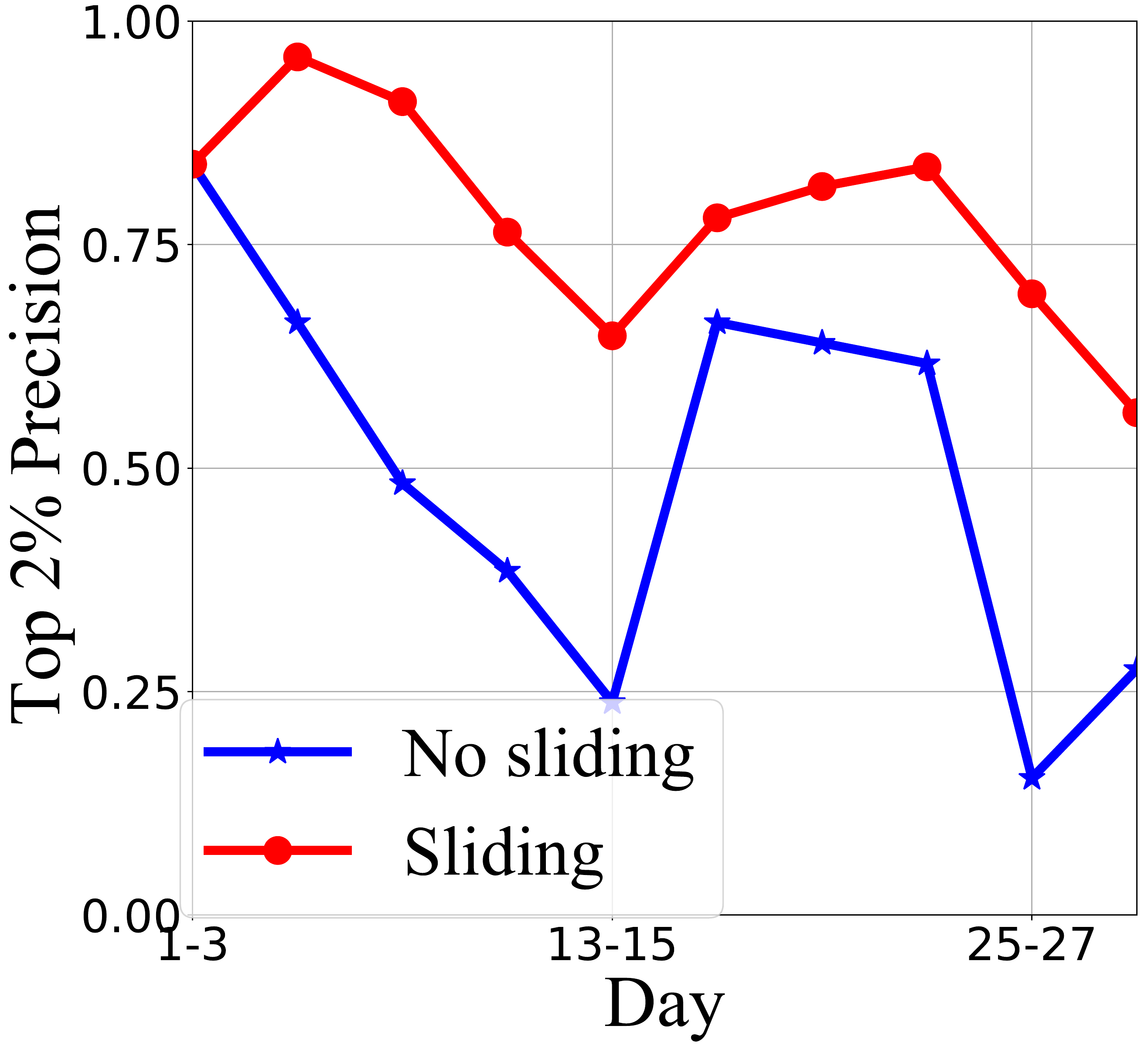}
     \label{fig:cas_slide}
     }
     \caption{Precision@K of parallel model and cascade model with and without sliding training on data in April.}
     \label{fig:slide_train}
 \end{figure}

To evaluate how much the sliding training helps improve the stability of precision, we test the daily precision@K.
Since the retrain period $T^{retrain}$ is set to be 3 days, we calculate an average precision@K for every 3 days.
Figure~\ref{fig:slide_train} shows the average precision@K of parallel model and cascade model with and without sliding training.
From the figure we see the precision@K of models with sliding training is much more stable than the ones without it. With sliding training, the variance of precision@K decreases from 0.058 to 0.009 for parallel model, and from 0.051 to 0.014 for cascade model, which validates that the sliding training improves the precision stability.

\begin{table}[tb]
  \centering
   \scalebox{1.0}{
  \begin{tabular}{l c c c}
  \hline 
    Model & Precision@K & Recall@K & Accuracy  \\ \hline 
    1D-CNN & 67.0\% & 12.1\% & 89.4\% \\ 
    MLP & 66.7\% & 12.1\% & 89.2\%  \\ 
    GBDT & 65.2\% & 13.4\% & 88.9\%  \\ \hline 
    Parallel & 80.0\% & 10.0\% & 89.8\%  \\ 
    Cascade & 78.1\% & 14.1\% & 90.3\%  \\ \hline   
  \end{tabular}
  }
  \vspace{-0.1 in}
    \caption{Comparison of parallel model, cascade model, 1D-CNN. MLP and GBDT, all with sliding training on data in April.}
  \label{tab:ensemble_perf_slide}
\end{table}

We further compare the performance of parallel and cascade models against baselines, all with sliding training.
The results are shown in Table~\ref{tab:ensemble_perf_slide}. From the table we observe that compared to the best baseline model (i.e., 1D-CNN), the parallel model improves the precision@K by 13.0\% (from 67\% to 80.0\%), and the cascade model improves the precision@K by 11.1\% (from 67\% to 78.1\%).
Similar improvements are achieved on data collected in May and June, which confirms that the parallel and cascade model still outperforms baselines with sliding training.

\subsection{Evaluation of Variable-Length Sliding Training}

\begin{figure}[t]
     \centering
     \subfigure[Parallel model.]{
         \includegraphics[width=1.57in]{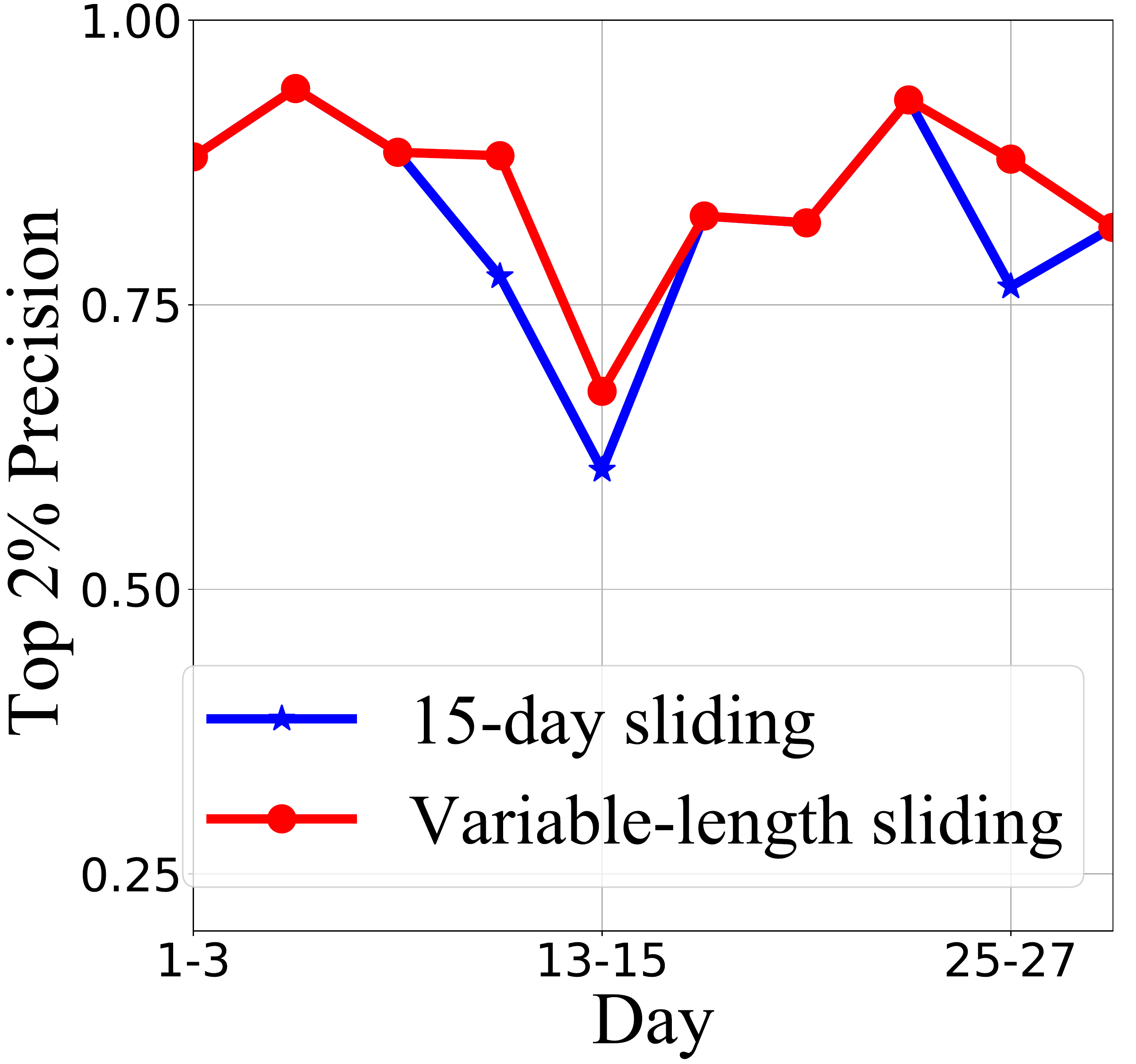}
     \label{fig:para_var_april}
     }
     \hspace{-0.1 in}
     \subfigure[Cascade model.]{
         \includegraphics[width=1.57in]{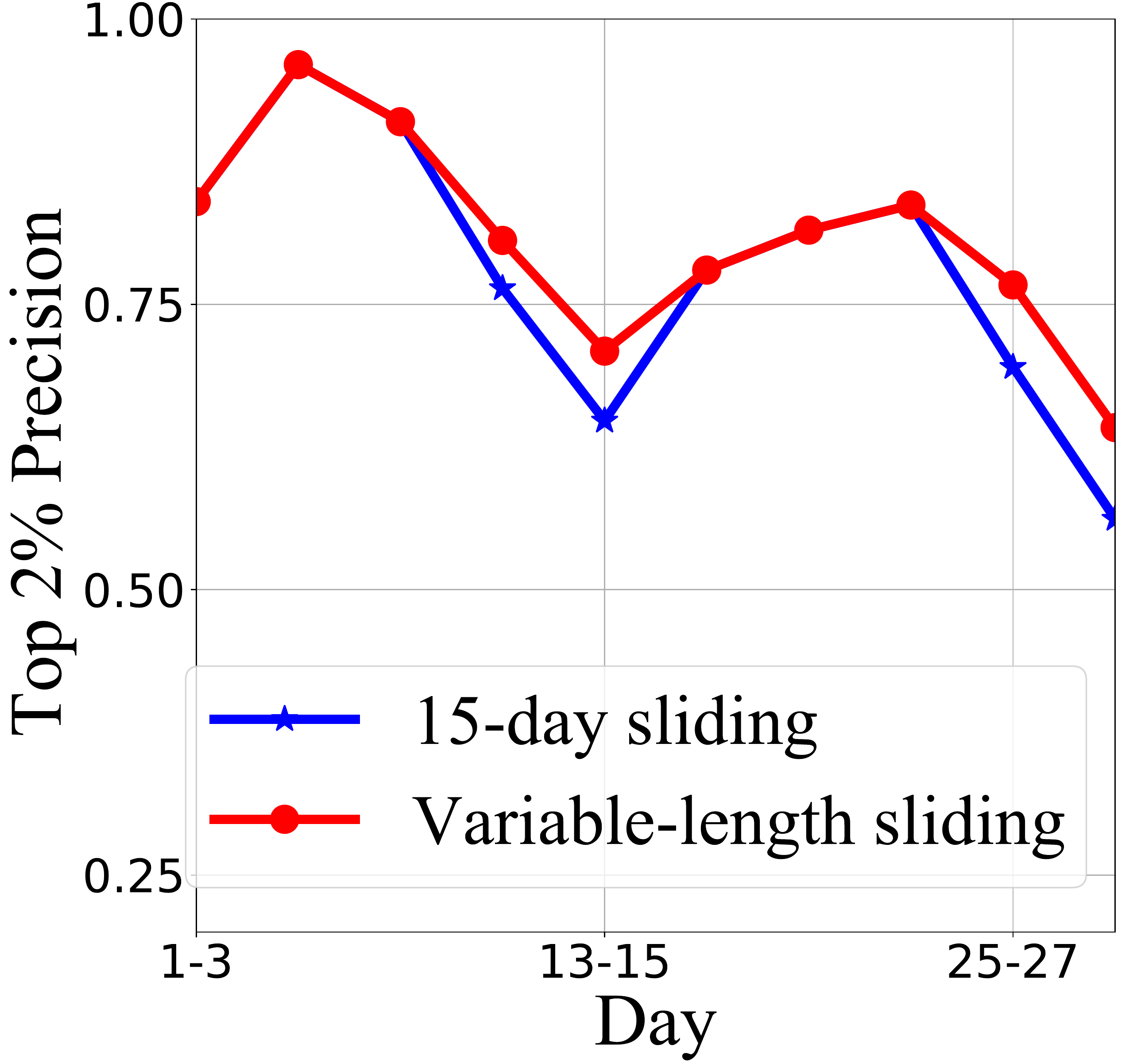}
     \label{fig:cas_var_april}
     } 
     \caption{Precision@K comparison of parallel model and cascade model with fixed-length and variable-length sliding training on data in April.}
     \label{fig:variable_april}
 \end{figure}

\begin{table}[t]
  \centering
  \begin{tabular}{l c c c}
    \hline
    Model  & Precision@K & Recall@K & Accuracy   \\ \hline
    Parallel (FL) & 80.0\% & 10.8\% & 90.2\%  \\ 
    Parallel (VL) & 85.4\% & 13.2\% & 91.5\%  \\ 
    Cascade (FL) & 78.1\% & 15.3\% & 91.4\%  \\ 
    Cascade (VL) & 80.7\% & 17.4\% & 91.8\%  \\ \hline   
  \end{tabular}
  \vspace{-0.1in}
    \caption{Comparison of parallel model and cascade model with fixed-length (FL) and variable-length (VL) sliding training respectively, on data in April.}
  \label{tab:variable_april}
\end{table}

To validate the effectiveness of the variable-length sliding training, 
we evaluate the parallel model and the cascade model with fixed-length sliding training and variable-length sliding training, respectively.
For fixed-length sliding training, the training set length $L^{train}$ is set to be 15 days.
For variable-length sliding training, we train three models with $L^{train}$ to be 9 days, 12 days and 15 days respectively, and use the model with the highest precision@K for each test window.
Figure~\ref{fig:variable_april} shows the precision@K of parallel model and cascade model with fixed-length sliding training and variable-length sliding training on data in April.
Table~\ref{tab:variable_april} shows the average precision@K, recall@K, and accuracy over one month. With variable-length sliding training, the precision@K is improved by 5.4\% for the parallel model, and 2.6\% for the cascade model.

 \begin{figure}[t!]
     \centering
      \subfigure[Parallel model.]{
         \includegraphics[width=1.57in, height=1.48in]{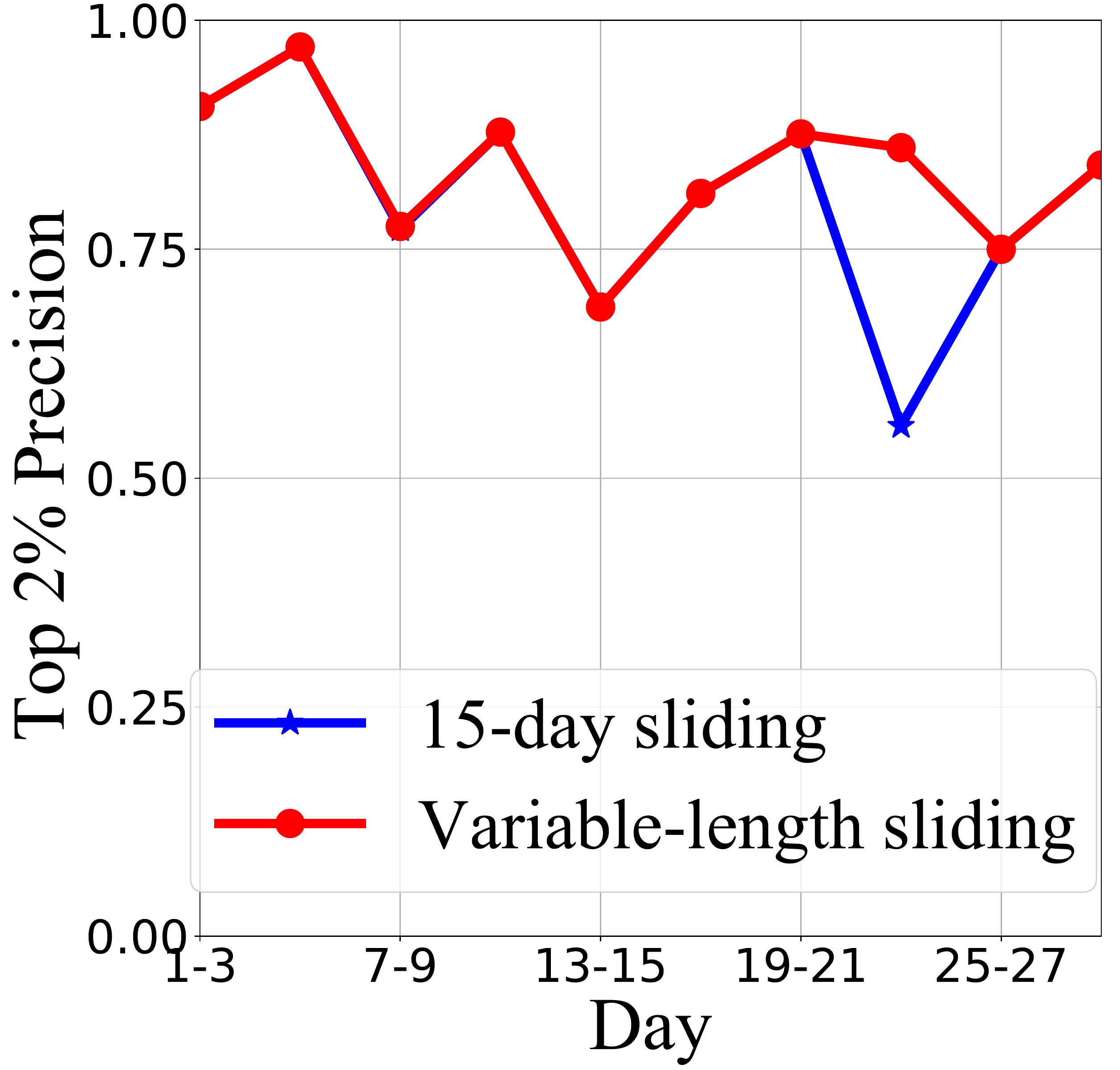}
     \label{fig:para_var_may}
     }
     \hspace{-0.1 in} 
     \subfigure[Cascade model.]{
         \includegraphics[width=1.57in, height=1.48in]{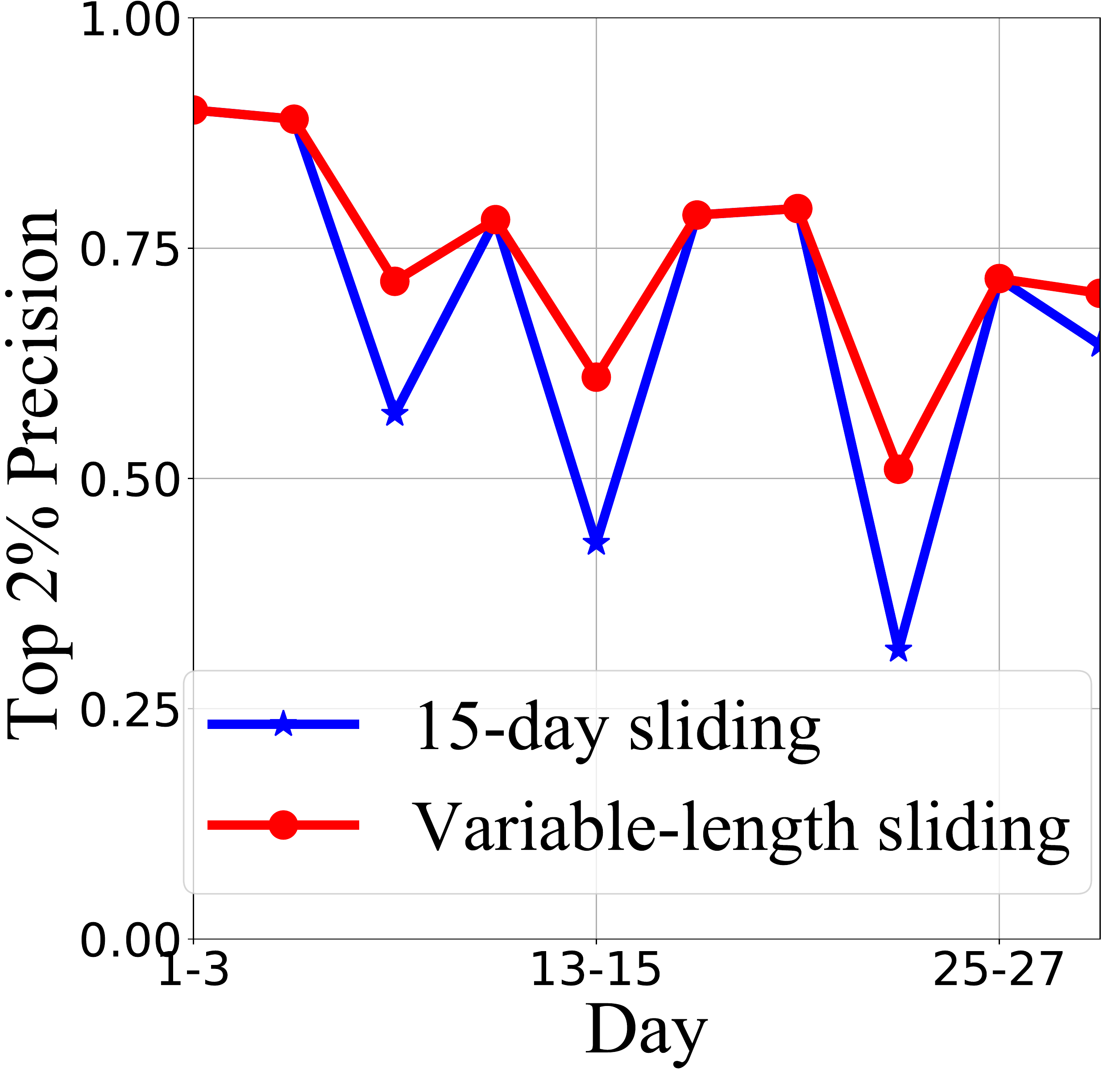}
     \label{fig:cas_var_may}
     }
     \caption{Precision@K comparison of parallel model and cascade model with fixed-length and variable-length sliding training on data in May.}
     \label{fig:variable_may}
 \end{figure}
 
 \begin{table}[t]
  \centering
  \begin{tabular}{l c c c}
    \hline
    Model  & Precision@K & Recall@K & Accuracy   \\ \hline 
    Parallel (FL) & 80.5\% & 10.9\% & 90.5\%  \\
    Parallel (VL) & 84.4\% & 12.7\% & 91.3\%  \\ 
    Cascade (FL) & 68.3\% & 12.1\% & 89.8\%  \\ 
    Cascade (VL) & 74.0\% & 14.5\% & 90.6\%  \\  \hline 
  \end{tabular}
  \vspace{-0.1in}
    \caption{Comparison of parallel model and cascade model with fixed-length (FL) and variable-length (VL) sliding training on data in May.}
  \label{tab:variable_slide}
\end{table}

 \begin{figure}[t!]
     \centering
     \subfigure[Parallel model.]{
         \includegraphics[width=1.57in]{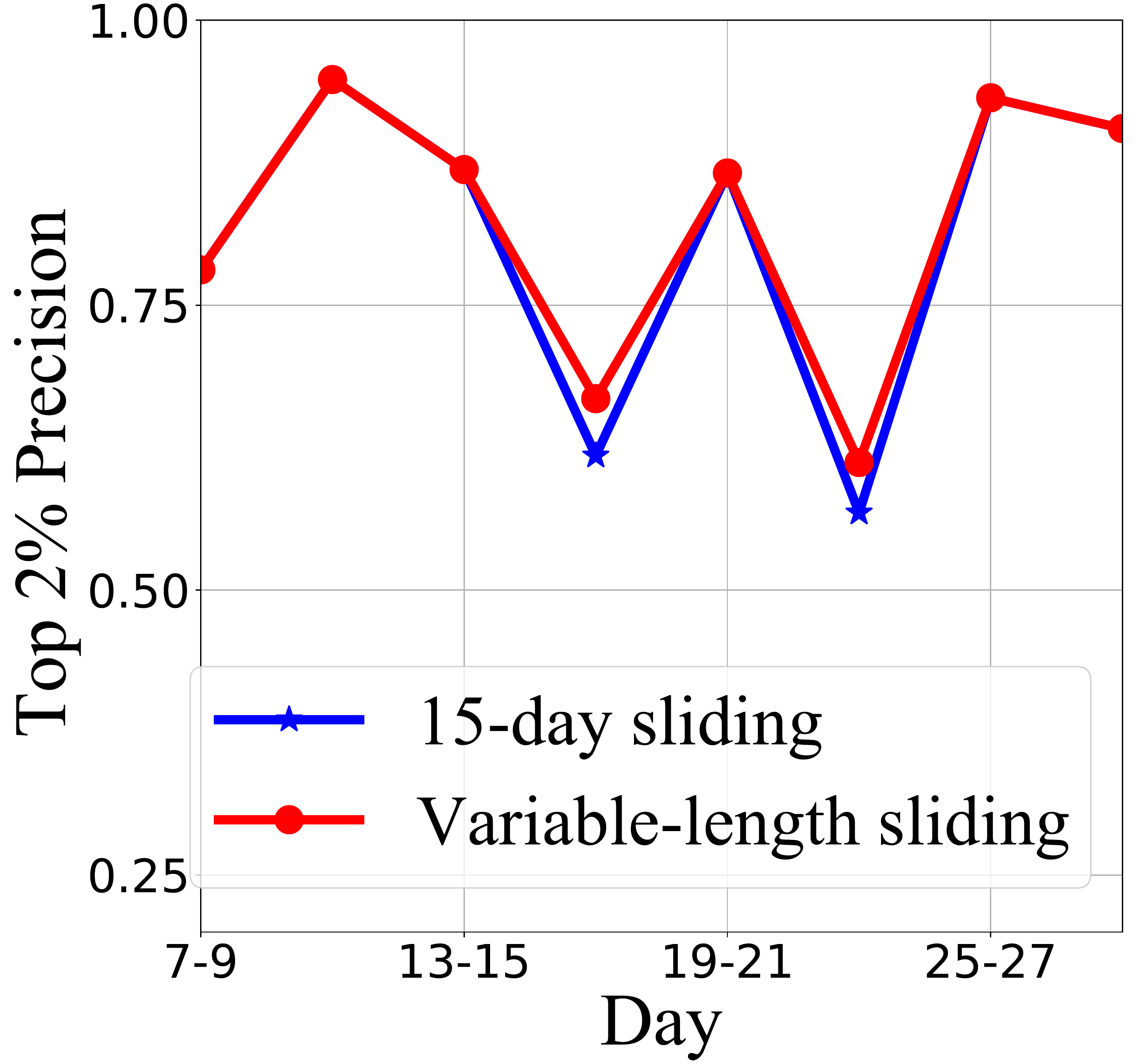}
     \label{fig:para_var_june}
     }
     \hspace{-0.1 in}
     \subfigure[Cascade model.]{
         \includegraphics[width=1.57in]{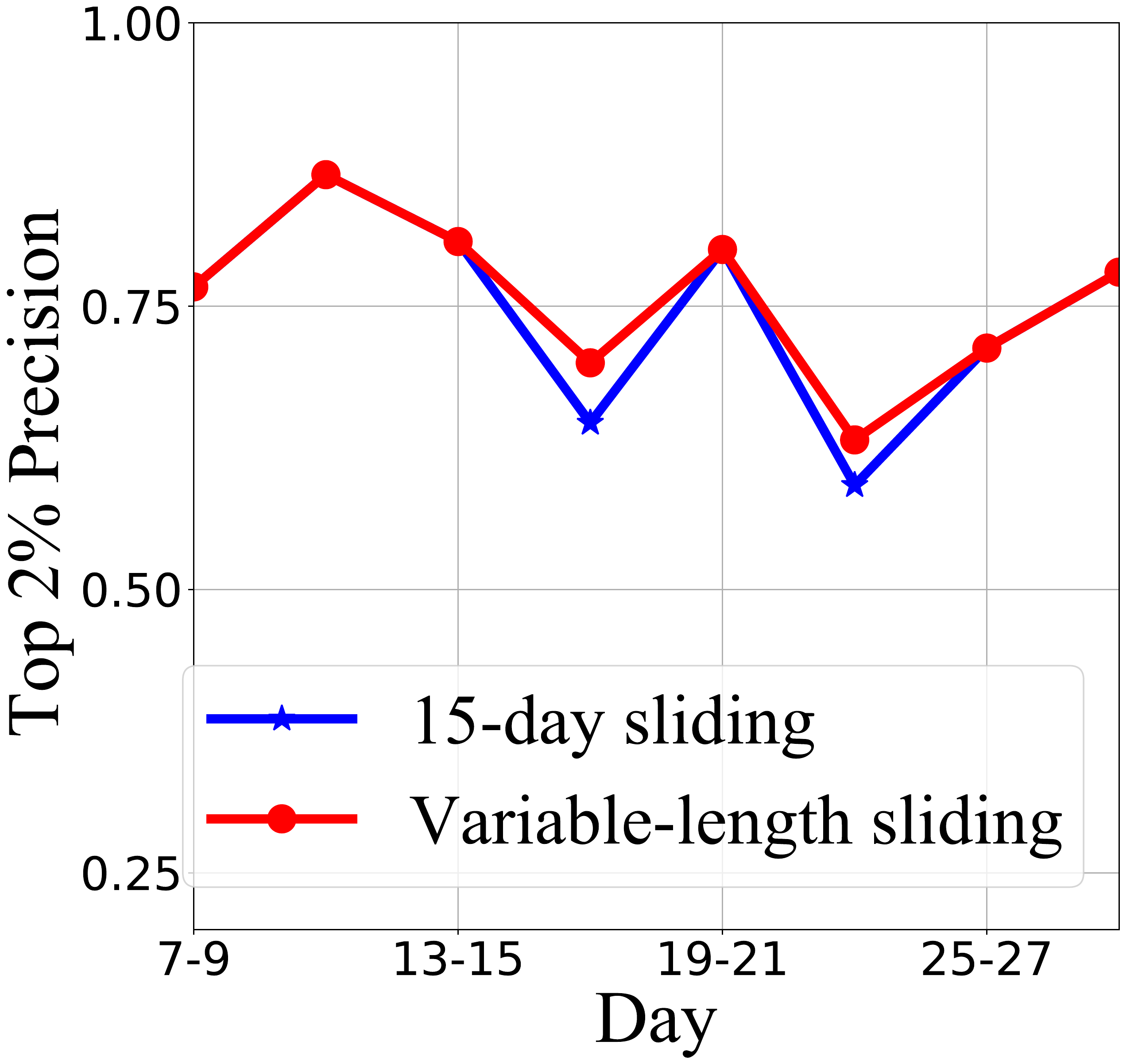}
     \label{fig:cas_var_june}
     } 
     \caption{Precision@K comparison of parallel model and cascade model with fixed-length and variable-length sliding training on data in June.}
     \label{fig:variable_june}
     \vspace{-0.5cm}
 \end{figure}

\begin{table}[t]
  \centering
  \begin{tabular}{l c c c}
    \hline
    Model  & Precision@K & Recall@K & Accuracy   \\ \hline 
    Parallel (FL) &  81.0\% & 11.7\% & 90.8\%  \\ 
    Parallel (VL) & 81.6\% & 12.0\% & 91.0\%  \\ 
    Cascade (FL) & 74.6\% & 14.8\% & 90.4\%  \\ 
    Cascade (VL) & 75.8\% & 14.9\% & 90.8\%  \\ \hline   
  \end{tabular}
  \vspace{-0.1in}
    \caption{Comparison of parallel model and cascade model with fixed-length (FL) and variable-length (VL) sliding training on data in June.}
  \label{tab:variable_slide}
\end{table}

To validate the generality of our proposed techniques we further present the model performance in May and June.
Figure~\ref{fig:variable_may} and Figure~\ref{fig:variable_june} show the performance of the parallel model and the cascade model with fixed-length sliding training and variable-length sliding training on the valid data collected in May and June. 
From the table we see that the parallel and cascade model with variable-length sliding training achieve precision@K of 84.4\% and 74.0\% in May, and 81.6\% and 78.5\% in June, 
which outperform the fixed-length sliding training, consistently. 
The average precisions of the parallel model and the cascade model over three months are 84.0\% and 76.9\%, respectively.

\section{Related Work}
\label{sec:relatedwork}

As stated earlier in Section~\ref{sec:introduction},
our study is the first to predict GPU failures in a large-scale DL cluster.
In this section, we list and discuss the most relevant works, organized in topics.

\textbf{Predicting failures.}
It is natural to consider leveraging models to predict failures.
Indeed, many prior works~\cite{liang2006bluegene,xue2015practise,botezatu2016predicting,gao2020task,das2018desh}
build algorithms and models to anticipate emerging failures.
Specifically, Liang et al.~\cite{liang2006bluegene} predict system failures by three heuristics
relating to failures' temporal characteristics, spatial characteristics, and
non-fatal events.
Similarly, Bontezatu et al.~\cite{botezatu2016predicting} build a classification model to
predict disk replacements using SMART attributes,
a set of sensor parameters for hard drives.

Beyond heuristic algorithms and classic statistic models,
neural networks are also used to predict assorted status in large-scale
systems.
PRACTISE~\cite{botezatu2016predicting}
is a time series prediction model based on neural networks
that forecasts future loads in a datacenter.
Though not aiming at failures, the idea of using neural networks is inspiring.
Gao et al.~\cite{gao2020task} build deep neural networks to predict task failures in could data centers.
Desh~\cite{das2018desh} is another example of using neural networks
to predict system health for HPC.
The most relevant work~\cite{nie2018machine}
studies four machine learning models---neural networks included---to predict
single-bit errors in GPU memory.
Besides various differences in the context (e.g., different training features, HPC vs. DL workloads),
our approach focuses on building ML models to predict future GPU failures.
In addition, our method differs in the data preprocessing,
where we have designed several specialized methods to better assist model training.

\textbf{GPU failures.}
A line of research~\cite{tiwari2015understanding, nie2016large, nie2017characterizing, nie2018machine, gupta2015understanding}
on Titan supercomputer studies various GPU failures,
including investigating GPU errors in general~\cite{tiwari2015understanding},
analyzing GPU software errors~\cite{nie2016large},
and characterizing GPU failures with temperature and power~\cite{nie2017characterizing},
and failures' spatial characteristics~\cite{gupta2015understanding}.
A study~\cite{di2014lessons} about another supercomputer, Blue Water,
analyzes GPU failures among other hardware failures.
One of their observations is that GPUs are among the top-3 most failed
hardware and GPU memory is more sensitive to uncorrectable errors than main memory.
This highlights that compared with other hardware components,
GPUs are prone to failures.
On contract, we study the GPU failure prediction instead.

\textbf{Hardware failures.}
Besides GPUs,
there are studies focusing on failures of other hardware components,
including DRAM~\cite{sridharan2013feng, meza2015revisiting},
disks~\cite{pinheiro2007failure},
SSDs~\cite{schroeder2016flash},
co-processors like Xeon Phi~\cite{oliveira2017experimental},
and other datacenter hardware~\cite{vishwanath2010characterizing}.
Our study on GPU failure prediction can potentially be extended to other
hardware components as well.

\section{Discussion and Conclusion}
\label{sec:discuss}

In this section, we discuss our next steps of predicting GPU failures.
First, we plan to explore other forms of ensemble mechanisms, 
including combining cascade and parallel mechanisms,
and changing model numbers in the cascade and parallel mechanisms.
To combine cascade and parallel mechanisms,
we can replace the first (or the second) model in the cascade mechanism with a
parallel model.
%
Besides, there are several factors (e.g., number of models) that trade off the precision and recall in the ensemble mechanisms.
For the parallel mechanism, precision will improve when adding more models,
but the recall will decrease because fewer instances are selected.
For the cascade mechanism, precision will improve when we make the first model filter out more instances, but the recall will decrease because some positive instances may be filtered out. 

Second, we will explore how to automatically adjust the variable-length sliding training.
One possible solution is to integrate AutoML into the variable-length sliding training. 
Specifically, we can train several models with different training set lengths
and evaluate their performances on the recently collected data. Then the training
set length corresponds to the top-precision model will be used for the
training of the current model.


\textbf{Conclusion.}
Studying the models for GPU failure prediction is crucial, since GPU failures are common, expensive, and may lead to severe consequences in today’s large-scale deep learning clusters. 
This paper is the first to study the prediction of GPU failures under production-scale logs. 
We observe the challenges of GPU failure prediction, and propose several techniques to improve the precision and stability of the prediction models. 
The proposed techniques can also be used in other failure prediction problems, such as the failures prediction in DRAM, disks and SSDs.

\bibliography{paper}
\bibliographystyle{mlsys2022}

\end{document}